\documentclass[final]{cvpr}

\usepackage{times}
\usepackage{epsfig}
\usepackage{graphicx}
\usepackage{amsmath}
\usepackage{amssymb}

\usepackage{comment}
\usepackage{multirow}
\usepackage{algorithm,algpseudocode}
\usepackage{url}

\usepackage{bbding}
\usepackage{verbatim}%dongzhuoyao
\usepackage{wasysym} %dongzhuoyao
\usepackage{xspace}
\usepackage{tabularx}
\usepackage{sidecap}

\usepackage{booktabs}  % table
\usepackage[table,x11names]{xcolor}
\definecolor{mygray}{gray}{0.95}
\definecolor{myred}{rgb}{1.0, 0.0, 0.0}

\usepackage{caption} 
\makeatletter
\DeclareRobustCommand\onedot{\futurelet\@let@token\@onedot}
\def\@onedot{\ifx\@let@token.\else.\null\fi\xspace}

\def\eg{\emph{e.g}\onedot} 
\def\ie{\emph{i.e}\onedot} 
 
\def\etc{\emph{etc}\onedot} 
\def\wrt{w.r.t\onedot} 
\def\etal{\emph{et al}\onedot}
\makeatother

\usepackage{threeparttable}
\usepackage{microtype}
\usepackage{cite}
\usepackage[pagebackref=true,breaklinks=true,colorlinks,bookmarks=false]{hyperref}

\def\cvprPaperID{534} % *** Enter the CVPR Paper ID here
\def\confYear{CVPR 2021}

\begin{document}

%%%%%%%%% TITLE
% \title{Few-Shot Common Action Transformation into Time and Space}
\title{Few-Shot Transformation of Common Actions into Time and Space}

\author{Pengwan Yang\thanks{yangpengwan2016@gmail.com}, Pascal Mettes, Cees G. M. Snoek\\
University of Amsterdam

}

\maketitle

%%%%%%%%% ABSTRACT
\begin{abstract}
This paper introduces the task of few-shot common action localization in time and space. Given a few trimmed support videos containing the same but unknown action, we strive for spatio-temporal localization of that action in a long untrimmed query video. We do not require any class labels, interval bounds, or bounding boxes. To address this challenging task, we introduce a novel few-shot transformer architecture with a dedicated encoder-decoder structure optimized for joint commonality learning and localization prediction, without the need for proposals.
%An encoder processes spatio-temporal sequences of the few support videos and the untrimmed query video, while a decoder integrates the support branch into the query branch and aggregates the fused features into a final prediction. 
Experiments on our reorganizations of the AVA and UCF101-24 datasets show the effectiveness of our approach for few-shot common action localization, even when the support videos are noisy. Although we are not specifically designed for common localization in time only, we also compare favorably against the few-shot and one-shot state-of-the-art in this setting. Lastly, we demonstrate that the few-shot transformer is easily extended to common action localization per pixel.

\end{abstract}

%%%%%%%%% BODY TEXT

%=============================================
\section{Introduction}
\begin{figure}[t!]
	\centering
	\includegraphics[width=0.95\columnwidth]{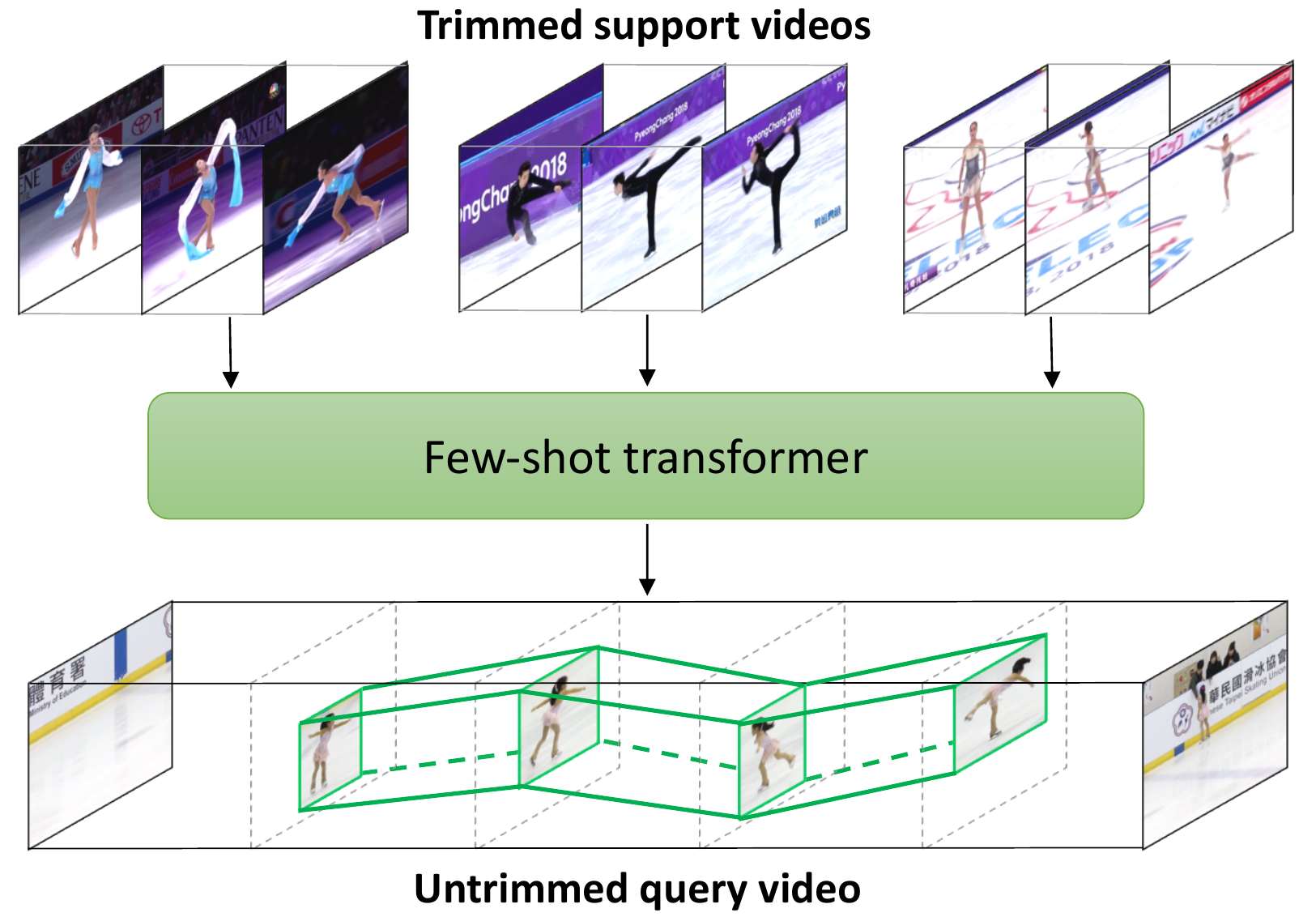}
	\caption{\textbf{Few-shot common action localization in time and space.} Given a few trimmed support videos sharing a common action, our proposed few-shot transformer is able to localize the spatio-temporal tubelet of the common action in a long untrimmed query video, without requiring the action class label, or any temporal or spatial annotations.}
	\label{fig:task}
\end{figure}

%=============================================
%Paragraph 1: goal and novelty
The goal of this paper is to localize an action in video time and space, without the need for class labels, interval bounds or box annotations. Class-agnostic action proposals, for either temporal, \eg~\cite{HeilbronCVPR16, ShouCVPR16, lin2019bmn} or spatio-temporal, \eg,~\cite{JainCVPR14, YuCVPR15, zhu2017tornado} action localization, have the same goal. However, to be effective they need to generate many proposals and a secondary \textit{supervised} step to find the best fitting one. To avoid the need for extensive supervision, Feng \etal \cite{feng2019spatio} introduced \textit{one-shot} localization of actions in time and space. They rely on proposals as well, but rather than using class supervision, a matching model between a trimmed support video and a long untrimmed video determines the best proposal. In similar spirit, Yang \etal \cite{yang2020localizing} introduced common action localization in \textit{time}. Given a few trimmed support videos containing the same (unknown) action, they are able to localize with the aid of proposals an action in a long untrimmed video. Their setup also avoids the need for temporal and class annotations.
In this work, we extend upon both \cite{feng2019spatio} and \cite{yang2020localizing} and propose the new task of \textit{few-shot} common action localization in \textit{time and space}. Our approach does not require any box annotations or class labels to obtain the spatio-temporal localization, and neither do we need proposals as in \cite{feng2019spatio,yang2020localizing}. All we require are a handful of trimmed videos showing a common unnamed action, see Figure~\ref{fig:task}. 

Our approach to few-shot common action localization is inspired by the success of transformers~\cite{bello2019attention} in object detection as demonstrated by Carion~\etal~\cite{carion2020detr}. Their approach eliminates the need for proposals and the accompanying hand-crafted components, while maintaining competitive performance. Such a proposal-free method avoids the needle-in-the-haystack problem with proposals in object detection, a problem which is even more severe in spatio-temporal action localization. Further, they exploit the versatile and powerful relation modeling capability of transformers. This is naturally suitable for common action localization as well, where the core challenge is to model the commonality between the few support videos and the single query video. To that end, we propose a transformer with an encoder-decoder structure that is adapted for our task of few-shot common action localization in time and space.

We make three contributions. First, we introduce the task of few-shot common action localization in time and space. We localize the spatio-temporal tube encapsulating an unknown action, based only on the commonality between a long untrimmed query video and a few trimmed support videos containing the same unknown action. Second, we propose a few-shot transformer with a dedicated encoder-decoder structure optimized for joint commonality learning and localization prediction, without the need for proposals. As a third contribution, we reorganize the videos in AVA~\cite{gu2018ava} and UCF101-24~\cite{soomro2012ucf101} to allow for evaluation of few-shot common action localization in time and space. Our experiments show the effectiveness of our approach, even when support videos are noisy. Moreover, we demonstrate  compatibility to few-shot and one-shot temporal action localization, outperforming the respective state-of-the-art. Last but not least, we show that the few-shot transformer easily extends to common action localization per pixel. %Code: \url{https://github.com/PengWan-Yang/few-shot-transformer}.
\section{Related work}
%=============================================

\textbf{Few- and zero-shot action localization.} 
Yang~\etal~\cite{yang2018one} introduce few-shot action-class localization in time, where a few (or at least one) positive labeled and several negative labeled videos steer the localization via an end-to-end meta-learning strategy. The strategy relies on sliding windows to swipe over the untrimmed query video to generate fixed boundary proposals. Xu~\etal~\cite{xu2020revisiting} also temporally localize an action from a few positive labeled and several negative labeled videos. They adopt a region proposal network~\cite{ren2015faster} to produce proposals with flexible boundaries. Zhang~\etal~\cite{zhang2020metal} perform few-shot temporal action localization, where video-level annotations are needed.  They construct a multi-scale feature pyramid to
directly produce temporal features at variable scales. For spatio-temporal action localization, a number of works have investigated a zero-shot perspective by linking actions to relevant objects~\cite{jain2015objects2action,mettes2017spatial,kalogeiton2017joint,mettesIJCV21}, or by leveraging trimmed videos used for action classification~\cite{jain2020actionbytes}. Soomro~\etal~\cite{soomro2017unsupervised} tackle action localization in an unsupervised setting, where no annotations are provided at all. While zero-shot and unsupervised action localization show promise, current approaches are not competitive with few-shot alternatives, hence we focus on the few-shot setting. Rather then relying on a few positive and many negative action class labels, like \cite{yang2018one,xu2020revisiting}, our action localization approach does not require any predefined positive nor negative action class labels, neither does it need any temporal or spatial bounds. All we require is that the few trimmed support videos have the same action in \emph{common}. Moreover, we attempt to localize the common action both in time and in space.

\textbf{Localization by commonality.} 
Compared with few-shot localization, common localization further frees the need for labels. It localizes according to the commonality between the query and support inputs. Kang~\etal~\cite{kang2019few} and Hu~\etal~\cite{hu2019silco} detect the common object in an image by a few examples containing the same object, without the need to know their class name. They both adopt a one-stage detector architecture with a feature reweighting module. Also for common object detection, Fan~\etal~\cite{fan2020few} rely on the support-guided region proposal network in a two-stage approach to produce support-related proposals and match each proposal with support images. Video relocalization by Feng~\etal~\cite{feng2018video} introduces temporal action localization in an untrimmed query video from a \textit{single} unlabeled support video, along with a one-stage approach. Yang~\etal~\cite{yang2020localizing} propose few-shot temporal localization of the common action. While their two-stage architecture is more general and effective than~\cite{feng2018video}, it depends on proposals and several hand-crafted components, such as anchors in the proposal generation and non-maximum suppression in a post-processing step~\cite{yang2020localizing}. Compared to these works, we aim for common localization in time and space simultaneously.

Closest to our work is spatio-temporal video relocalization by Feng~\etal~\cite{feng2019spatio}, which extends their temporal video relocalization~\cite{feng2018video} to time and space from a \textit{single} support video. Feng~\etal~\cite{feng2019spatio} propose a warp LSTM to align spatio-temporal information between adjacent frames to embed long-term dependencies. They produce a large set of proposals for each query video clip to match each proposal with their single support video. Like Yang~\etal~\cite{yang2020localizing}, their method requires hand-designed components in the proposal generation and post-processing step. Compared to spatio-temporal video relocalization, we consider a more general and realistic setting, where more than one support video can be used. We propose a transformer network architecture, which avoids the need for proposals. Lastly, we also generalize to common temporal and pixel-level action localization.
%and other pre-defined components. 

\begin{figure*}[th!]
	\centering
	\includegraphics[width=0.99\linewidth]{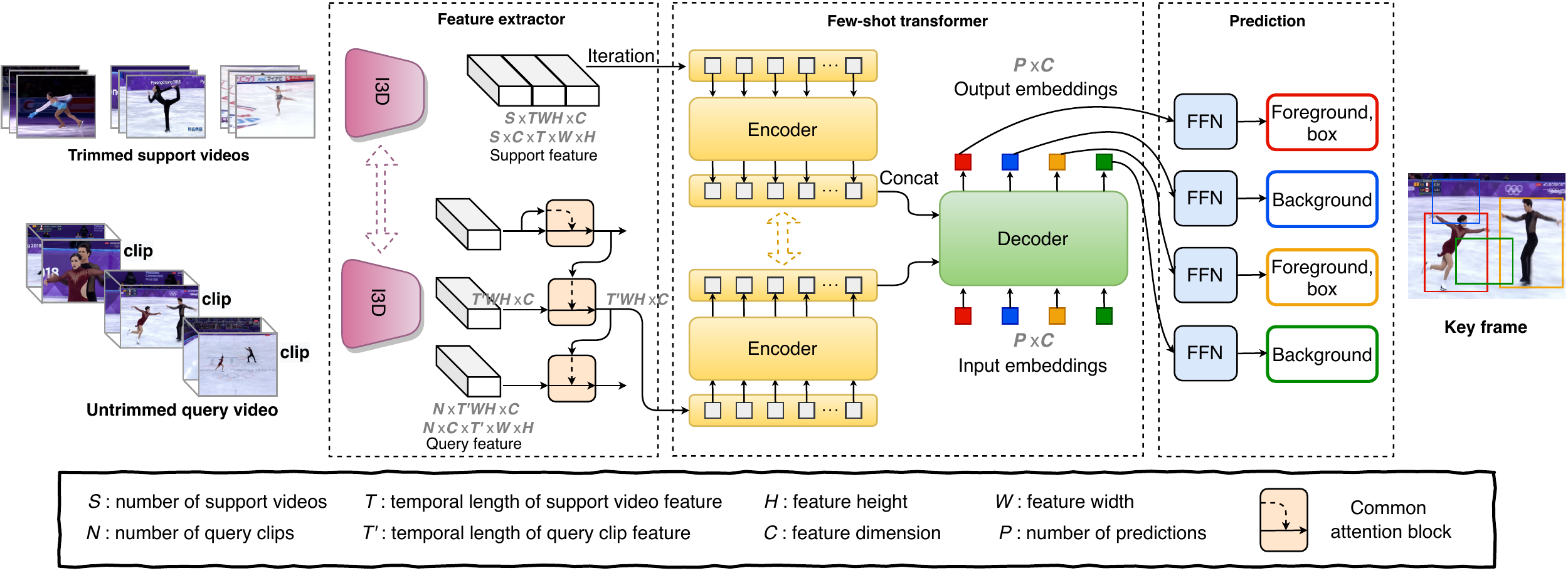}
	\caption{\textbf{Overview of the proposed model.} In the feature extraction stage, we obtain the representations of the support videos and the query video clips, and align each query clip feature with previous clips to contain more motion information. In the few-shot transformer, we perform further processing and fusion on the support features and query clip feature. Then we transform the input embeddings to output embeddings by aggregating it with the fused feature. Finally, the output embeddings are used for prediction in the prediction network, which contains a $3$-layer feed-forward neural network (FFN) and a linear projection. The detailed structure of the few-shot transformer is depicted in Figure~\ref{fig:module}. } 
	\label{fig:architecture}
\end{figure*}

\textbf{Transformers in vision.}
Transformers were introduced by Vaswani~\etal~\cite{vaswani2017attention} as a powerful attention-based building block for machine translation. Jaderberg~\etal~\cite{jaderberg2015spatial} proposed to spatially transform image feature maps, conditional on the feature map itself, alike the self-attention mechanism, even before the standard transformer structure became popular. Gavrilyuk~\etal~\cite{gavrilyuk2020actor} propose an actor-transformer to learn and selectively extract information relevant for group activity recognition. Ye~\etal~\cite{ye2020few} are the first to employ a self-attention mechanism via a transformer to contextualize instances in the few-shot image classification setting. 
Recently, Carion~\etal~\cite{carion2020detr} propose a transformer encoder-decoder structure for object detection to eliminate the need for many hand-designed components, while still demonstrating good performance. Encouraged by this early success of transformers for various vision challenges, we propose a transformer for few-shot common action localization based on the hypothesis that the transformer's attention mechanism can efficiently process sequences of spatio-temporal information and model the relation between query and support videos. Our one-stage transformer frees the need for proposals and other pre-defined components, typical in common action localization, while obtaining better localization results in time \textit{and} space. 
%Like Ye~\etal~\cite{ye2020few}, we also rely on the self-attention mechanism. Where Ye~\etal use self-attention to push support and query embeddings toward class prototypes for image classification, we exploit self-attention to highlight correspondences between an untrimmed query video and trimmed support videos, for common action detection in time and space. 

%=============================================
\section{Method}
%=============================================
Our goal is to localize the spatio-temporal tubelet of an action in an untrimmed query video based on the common action in the trimmed support videos. Note we have no access to the action class label or its time and space boundary annotations. Following recent work on \textit{supervised} spatio-temporal action detection~\eg,~\cite{gu2018ava,kalogeiton2017action,yang2019step}, our approach performs common localization at clip level,~\ie, common localization results are first obtained from each query video clip and then linked to build the action tubes across the whole query video.
Our approach consists of three main stages. First, we split the untrimmed query video into clips and feed them into the feature extractor, together with the few support videos, to obtain spatio-temporal representations. Each query clip feature is aligned with previous clip features using our common attention block. 
%\cs{Why is it called basic? Why is it mentioned here, why is the multi-head attention not mentioned? Are they important?}~\pengwan{For discussion: Basic block and multi-head block are both vital components of few-shot transformer. Besides, basic block can also play independent role, multi-head block can not. I mention basic block here, because it has an important role here. And in supplementary I will list results W/O  basic block at here. I want to mention multi-head block also, but it doesn't play independent role, I can't find a good angle to mention it. Maybe we should not mention basic block here also. And, maybe we can not claim the blocks alone as novelties. Only they combined with transformer structure can be claimed as novelties.} 
Then our few-shot transformer further processes the features, fuses the support features into the query clip feature, and transforms the input embeddings by aggregating it with the fused feature. Finally, a $3$-layer feed-forward neural network and a linear projection are added on top of the output embeddings from the few-shot transformer to make the final tubelet prediction. Our method is summarized in Figure~\ref{fig:architecture} and described for each stage in more detail next.

\begin{figure*}[t!]
	\centering
	\includegraphics[width=0.95\linewidth]{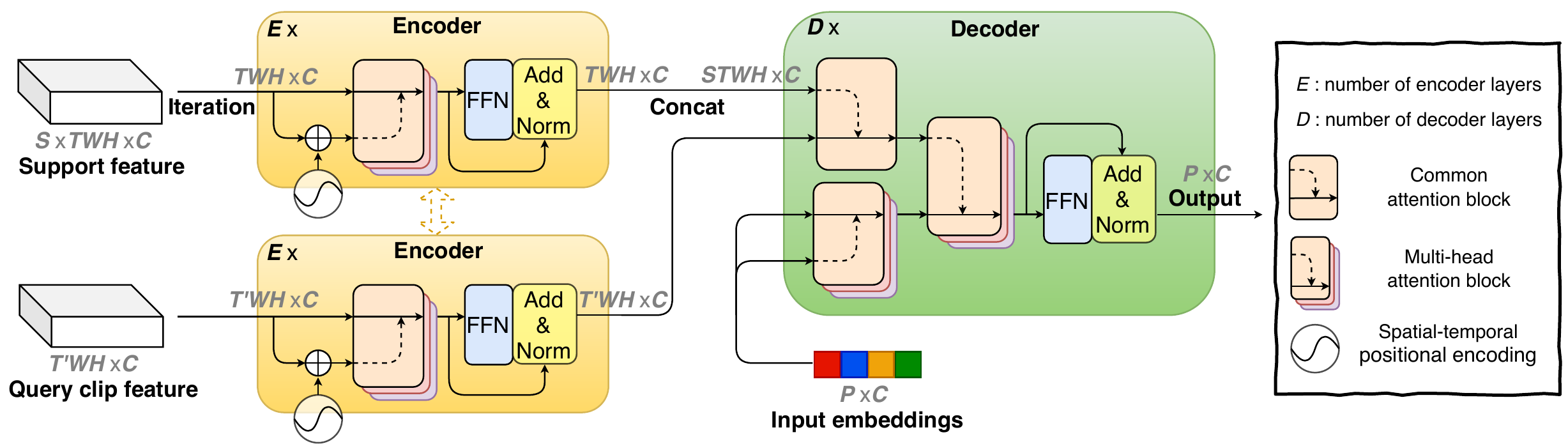}
	\caption{\textbf{Structure of few-shot transformer.} The encoder performs further processing on the support features and query clip feature. The decoder fuses the support feature into the query clip feature, then integrate the fused feature into the input embedings to obtain the output embeddings for the final prediction. }
	\label{fig:module}
\end{figure*}

%---------------------------------------------
\subsection{Video feature extractor}
%---------------------------------------------
We adopt an I3D network~\cite{carreira2017quo} as our backbone to obtain spatio-temporal representations of a single query video and a few support videos. The weights of the backbone network are shared between the support and query videos. For the support videos, we feed the whole video into the backbone directly, one by one.  Let $f^s \in \mathbb{R}^{S {\times} TWH {\times} C}$ denote the representation of all $S$ support videos with their spatio-temporal dimensions flattened, and $f^s_i \in  \mathbb{R}^{TWH {\times} C}$ denote the feature of the $i$-th support video. For a single untrimmed query video, we first split the video into multiple clips. Then the query clips go through the backbone network. Let $f^q \in \mathbb{R}^{N {\times} T'WH {\times} C}$ denote the representation of all $N$ query video clips, and $f^q_j \in \mathbb{R}^{ T'WH {\times} C}$ the representation of the $j$-th query clip. 

As is, the query clips are separate representations, \ie, $f^q_j$ only contains the spatio-temporal information within the $j$-th query video clip. To accurately localize the tubelets in the query video, long-term spatio-temporal information needs to be modeled. Our common attention block solves this problem. The common attention block is built on the self-attention mechanism~\cite{vaswani2017attention} and the non-local structure~\cite{wang2018non}. The self-attention mechanism in the transformer is extended to self- and cross-attention $\mathbb{A}$:
\begin{equation}
% \begin{split}
\mathbb{A}(\mathit{I}_1,\mathit{I}_2)=\text{softmax}(\frac{Q(\mathit{I}_1)K(\mathit{I}_2)^T}{\sqrt{C}})V(\mathit{I}_2),
\label{equ:attention}
% \end{split}
\end{equation}
where $Q(\cdot)$,$K(\cdot)$,$V(\cdot)$ are linear projections. $\mathit{I}_1$,$\mathit{I}_2$ denote two inputs, both of which have dimension $C$. The common attention block $\widehat{\mathbb{A}}$ can be defined as:
\begin{equation}
% \begin{split}
\widehat{\mathbb{A}}(\mathit{I}_1,\mathit{I}_2) = \mathit{I}_1 + \text{Linear}(\text{LayerNorm}(\mathbb{A}(\mathit{I}_1,\mathit{I}_2))),
\label{equ:clip}
% \end{split}
\end{equation}
where for simplicity, $\text{Dropout}$~\cite{srivastava2014dropout} and $\text{ReLU}$~\cite{glorot2011deep} operations are ignored here. A detailed overview and illustration of the common attention block is provided in the supplementary materials. As demonstrated by prior work~\cite{hu2019silco,yang2020localizing,liu2020crnet}, a self- and cross-attention mechanism can efficiently enhance a representation by itself or by other representations. In our model, the common attention block plays two important roles. In the feature extraction stage, the common attention block aligns each query clip feature with its previous clip features, making each query clip feature contain more motion information, which benefits the common action localization. In the few-shot transformer structure, the common attention block fuses the support feature into the query clip feature, making common action localization achievable.

We utilize the common attention block $\widehat{\mathbb{A}}$ to propagate the spatio-temporal information from previous clips of the query video into the $j$-th query clip for better common localization:
\begin{equation}
% \begin{split}
\hat{f}^q_j=\widehat{\mathbb{A}}(f^q_j, \hat{f}^q_{j-1}),
\label{equ:clip_aligh}
% \end{split}
\end{equation}
where $\hat{f}^q_j \in \mathbb{R}^{ T'WH {\times} C}$ is the $j$-th query clip representation enhanced by previous clips.

%---------------------------------------------
\subsection{Few-shot transformer}
%---------------------------------------------
In the standard transformer, and in our few-shot transformer as well (detailed in Figure~\ref{fig:module}), the multi-head attention block is a key element. The multi-head attention block $\widehat{\mathbb{MA}}$ is built on multi-head attention $\mathbb{MA}$ which is an extension of attention $\mathbb{A}$:
\begin{equation}
% \begin{split}
\mathbb{MA}(\mathit{I}_1,\mathit{I}_2)=\text{concat}(\mathbb{A}_1,\cdots,\mathbb{A}_m)W,
\label{equ:multi_attention}
% \end{split}
\end{equation}
\begin{equation}
% \begin{split}
\widehat{\mathbb{MA}}(\mathit{I}_1,\mathit{I}_2)=\text{LayerNorm}(\mathit{I}_1 + \mathbb{MA}(\mathit{I}_1,\mathit{I}_2)),
\label{equ:multi_attention_block}
% \end{split}
\end{equation}
where $m$ denotes the number of heads, $W$ denotes a linear projection, $\mathbb{A}_1,\cdots,\mathbb{A}_m$ are parallel instances of $\mathbb{A}(\mathit{I}_1,\mathit{I}_2)$.

\textbf{Encoder.}
Each encoder layer has a standard architecture and consists of a multi-head attention block $\mathbb{MA}$ and a feed forward network, $\text{FFN}$. Since the encoder architecture is permutation-invariant, we supplement it with fixed spatio-temporal positional encodings~\cite{bello2019attention,parmar2018image} that are added to both support video and query clip features in each encoder layer. For the support branch, we let the support video go through the encoder one by one, then concatenate the output and get a support branch input $E^s\in \mathbb{R}^{STWH {\times} C}$ to the decoder:
%
% \begin{equation}
% L(X)=\text{Linear}(\text{Dropout}(\text{ReLU}(\text{Linear}(X))))
% \label{equ:ffn}
% \end{equation}
%
\begin{equation}
\hat{E}(f^s_i)=\widehat{\mathbb{MA}}(f^s_i, \textit{pos}(f^s_i)),
\label{equ:e_S_i}
\end{equation}
\begin{equation}
E(f^s_i)=\text{LayerNorm}(\hat{E}(f^s_i)+\text{FFN}(\hat{E}(f^s_i))),
\label{equ:E_S_i}
\end{equation}
\begin{equation}
E^s=\text{concat}(E(f^s_1),\cdots,E(f^s_S)),
\label{equ:E_S}
\end{equation}
where $S$ denotes the number of support videos and $\textit{pos}(\cdot)$ denotes the operation of adding spatio-temporal positional encoding, FFN can be seen as a $1 {\times} 1$ convolutional layer. For the query branch, we let the enhanced query clip feature $\hat{f^q_j}$ go through the encoder and get the query branch input to the decoder $E^q{=}E(\hat{f}^q_j)\in \mathbb{R}^{T'WH {\times} C}$. 

\textbf{Decoder.} In a standard transformer decoder there are two inputs, but in our few-shot transformer decoder there are three inputs: the support branch input $E^s$, the query branch input $E^q$ from the encoder, and the $input \ embeddings$, which are the learnt positional encodings. For the decoding, we first fuse the support feature into the query clip feature by using the common attention block,~\ie $\widehat{\mathbb{A}}(E^q,E^s)$. Then, we perform self-attention on the $input \ embeddings$ via the multi-head attention block,~\ie $\widehat{\mathbb{MA}}(\textit{in},\textit{in})$: 
\begin{equation}
\hat{D}=\widehat{\mathbb{MA}}(\widehat{\mathbb{A}}(E^q,E^s), \widehat{\mathbb{MA}}(\textit{in},\textit{in})).
\label{equ:decoder_1}
\end{equation}
The $input \ embeddings$ are transformed into $output \ embeddings$ by aggregating the fused feature through another multi-head attention block: 
\begin{equation}
\textit{out}=\text{LayerNorm}(\hat{D}+\text{FFN}(\hat{D})),
\label{equ:decoder_2}
\end{equation}
which we use as representation for our final prediction.

%---------------------------------------------
\subsection{Prediction network}
%---------------------------------------------
For final prediction, a 3-layer perceptron is performed on the $output \ embedding$ of the few-shot transformer, with a ReLU activation and hidden dimension $C$, followed by a linear set-based projection layer. The predicted results are the normalized center coordinates, the height and the width of the boxs~\wrt the keyframe of the $j-$th query clip, along with the $foreground/background$ binary labels. Just like Carion \etal~\cite{carion2020detr}, we use a set-based Hungarian loss~\cite{stewart2016end,kuhn1955hungarian} that forces unique predictions for each ground-truth bounding box via bipartite matching. Given the clip-level common localization results, we link them over time~\cite{kalogeiton2017action} to obtain the final action tubes for the whole untrimmed query video. 
%We follow the same linking algorithm as described in Kalogeiton~\etal~\cite{kalogeiton2017action}. 

%=============================================
\section{Experimental setup}
%=============================================
%---------------------------------------------
%\subsection{Datasets}
%---------------------------------------------

Existing video datasets are usually created for other vision tasks, such as classification~\cite{kay2017kinetics,idrees2017thumos}, temporal localization~\cite{caba2015activitynet}, action recognition~\cite{soomro2012ucf101}, captioning~\cite{chen2011collecting}, or summarization~\cite{gygli2014creating}. Feng~\etal~\cite{feng2019spatio} reorganize the videos in  AVA~\cite{gu2018ava} (version 2.1) into a new dataset for their one-shot spatio-temporal video relocalization task, which is unsuitable for the few-shot setting of our task. To evaluate few-shot spatio-temporal common action localization, we have revised two existing datasets that come with spatio-temporal annotations suitable for our evaluation, namely AVA~\cite{gu2018ava} (version 2.2) and UCF101-24~\cite{soomro2012ucf101}.

\textbf{Common-AVA.} There are 430 15-minute video clips with per-second action bounding box annotations in AVA~\cite{gu2018ava}. The annotated actions cover 80 atomic action categories, including ``stand",``talk", ``listen",~\etc. The actions are exhaustively annotated, covering 1.11 million action annotations with multiple labels per person. We link the consecutive bounding boxes if they have the same subject with all the action labels being the same. After linking, the tubelets with exactly the same action labels are regarded as semantically corresponding to each other. Hence, we combine multiple atomic action labels that are annotated with one bounding box together. Next, we discard the combined classes with less than 32 instances and obtain a total of 356 combined classes. We randomly select 80\% of the classes for training, 10\% of the classes for validation, and the remaining 10\% of the classes for testing. 

\textbf{Common-UCF.} UCF101~\cite{soomro2012ucf101} is originally an action classification dataset, and a subset of 24 classes with 3,207 videos are provided with  spatio-temporal annotations for action detection~\cite{soomro2012ucf101}. We use the recently revised ground truth action tube annotations from Singh \etal~\cite{singh2017online}. In UCF101-24, each action has a single class label. Among the 24 action categories in UCF101-24, 16 are used for training, 4 for validation and 4 for testing.

During training, the support videos and query video are randomly paired, while the pairs are fixed for validation and testing. Note that we do not use action class labels. More details of the two reorganized datasets are listed in Table~\ref{tab:dataset}.

\begin{table}[t] 
\centering
 \resizebox{1\columnwidth}{!}{
\begin{tabular}{lrr}
\toprule
 & \textbf{Common-AVA} & \textbf{Common-UCF} \\
\midrule
\rowcolor{mygray}
\textbf{Video statistics} & & \\
mean number of instances & 1 & 1.4 \\
mean length (sec) & 4.3 & 7.1 \\
frames per second & 25 & 25 \\
number of train videos & 160,889 & 2,418  \\
number of val+test videos & 12,794 & 776  \\
\midrule
\rowcolor{mygray}
\textbf{Class statistics} & & \\
number of train actions & 286 & 16  \\
number of val+test actions & 70 & 8  \\
\bottomrule
\end{tabular}
}
\caption{\textbf{Overview of the common spatio-temporal localization datasets.} Common-AVA contains a single target action per video, while Common-UCF contains videos with more action instances and longer duration.}
\label{tab:dataset}
\end{table}

%---------------------------------------------
%\subsection{Implementation and training details}
%---------------------------------------------
\textbf{Implementation details.} We resize all the videos to a 320 × 320 resolution before feeding them into the backbone. The I3D model we use is first initialized by training on the Kinetics dataset~\cite{kay2017kinetics} and then fine-tuned during the training of our model. We initialize the $input \ embeddings$ by setting the values to zeros and we set its dimensionality $P$ to 10. It determines the number of predicted boxes for each keyframe. The encoder and decoder both contain 6 layers. The length of the trimmed support video is 1 second. To form a batch during the training process, the length of the query videos needs to be fixed. The query video is fixed to be 4 seconds long by randomly cropping or padding zeros in the training process. During testing, the query video in full length is fed into the model without any batching and cropping. For Common-AVA the query clip is 1 second long, as a keyframe is annotated per second in the original AVA dataset. For Common-UCF, the query clip is $\frac{1}{5}$ second long because each frame in the videos is annotated at 25 FPS. So for Common-UCF, each query clip contains 5 frames, and we treat the middle frame as the keyframe of the clip.

\textbf{Training regime.} We train our model with AdamW~\cite{loshchilov2018decoupled}, setting the initial few-shot transformer’s learning rate to $ {10}^{-4} $, the backbone’s to $ {10}^{-5} $, and weight decay to $ {10}^{-4} $. We train for 100 epochs with a learning rate drop by a factor of 10 after 70 epochs, where a single epoch is a pass over 10K training videos for Common-AVA and all training videos for Common-UCF. The batch size is set to be 12 with four Nvidia GTX 1080TI cards. 

%---------------------------------------------
\textbf{Evaluation metrics.}
%---------------------------------------------
We report the frame-level mean average precision (frame-mAP) with an IoU threshold of 0.5 for both datasets. A prediction is correct when it has the correct $foreground/background$ prediction and has a ground truth overlap larger than the overlap threshold. On Common-UCF we also report the video-mAP.
%=============================================
\section{Results}
%=============================================
%------------------------------------------
%\subsection{Ablation study}
%------------------------------------------

\textbf{Few-shot transformer ablation.} We evaluate the effect of the encoder and decoder in our few-shot transformer. We report results for one and five support videos in Table~\ref{tab:module evaluation}. The decoder alone can achieve common spatio-temporal localization and obtains competitive performance. Performance improves notably when we add encoder processing of either support videos or query video. We can gain 5.3 mAP increase under the one-shot setting and 6.2 mAP increase under the five-shot setting, if we let both support and query features go through the encoder.

\textbf{Benefit of positional encoding.} We report results in Table~\ref{tab:positoinan encoding}. Including the spatio-temporal positional encoding is always beneficial, ideally in both the support and query encoder. Not passing any spatio-temporal positional encoding to the features leads to an mAP drop of 2.3 under the one-shot and an mAP drop of 3.6 under the five-shot setting.

\begin{figure}[tb!]
	\centering
	\includegraphics[width=0.95\columnwidth,trim=56 0 0 20,clip]{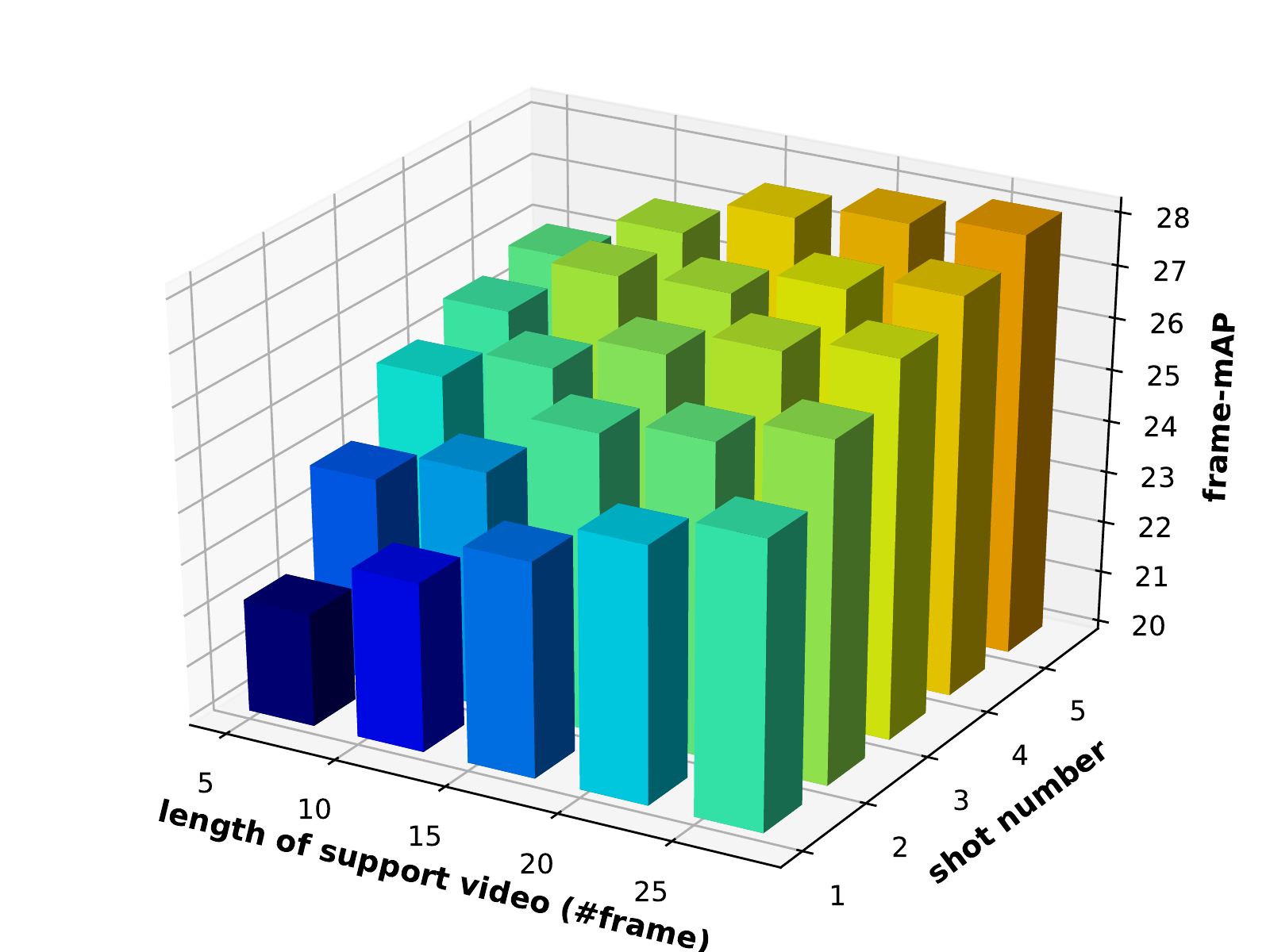}
	\caption{\textbf{Influence of length and number of support videos} on Common-AVA. We obtain a more precise common localization with more and longer support videos.}
	\label{fig:support_ablation}
\end{figure}

\textbf{Influence of support video length and amount.} Next, we ablate the effect of the length and the number of support videos on the spatio-temporal localization performance on Common-AVA in Figure~\ref{fig:support_ablation}. We sample 5, 10, 15, 20 and 25 frames for each support video and increase the support video number from 1 to 5. We find that the result gradually increases with longer support videos and/or more support videos, indicating that increasing the support information is beneficial to our modules for common action localization.

\begin{table}[t!]
	\centering
	\resizebox{0.95\columnwidth}{!}{%
	\begin{tabular}{ccccc}
	\toprule
	\multicolumn{2}{c}{\textbf{Encoder}} & \textbf{Decoder} & \multicolumn{2}{c}{\textbf{Few-shot setting}}\\
	\cmidrule(lr){1-2} 	\cmidrule(lr){3-3}	\cmidrule(lr){4-5}
	 support & query &  & one-shot & five-shot\\
	\midrule
	& & $\checkmark$ & 20.3 & 21.9\\
    & $\checkmark$ & $\checkmark$ & 23.8 & 25.0\\
    $\checkmark$ & & $\checkmark$ & 22.4 & 24.5\\
	$\checkmark$ & $\checkmark$ & $\checkmark$ & \textbf{25.6} & \textbf{28.1}\\
	\bottomrule
    \end{tabular}%
	}
	\caption{\textbf{Few-shot transformer ablation} on Common-AVA. The decoder achieves competitive performance by itself. It improves further when the encoder processes the support and/or query video. Best performance is obtained if the encoder processes both the support and query videos.}
	\label{tab:module evaluation}
\end{table}

\begin{table}
	\centering
	\resizebox{0.82\columnwidth}{!}{%
	\begin{tabular}{cccc}
	\toprule
	\multicolumn{2}{c}{\textbf{Positional encoding}} & \multicolumn{2}{c}{\textbf{Few-shot setting}}\\
	\cmidrule(lr){1-2} \cmidrule(lr){3-4}	
	support & query & one-shot & five-shot\\
	\midrule
	& & 23.3 & 24.5\\
    & $\checkmark$ & 24.4 & 25.8\\
    $\checkmark$ & & 24.1 & 26.4\\
	$\checkmark$ & $\checkmark$ & \textbf{25.6} & \textbf{28.1}\\
	\bottomrule
    \end{tabular}%
	}
	\caption{\textbf{Benefit of positional encoding.} Adding spatio-temporal positional encoding to the support or query features improves performance on Common-AVA. Including it in both support and query leads to best results. }
	\label{tab:positoinan encoding}
\end{table}

\textbf{Effect of noisy support videos.} To test the robustness of our approach, we ablate the effect of noisy inputs in the five-shot setting. Video-level noise is simulated by including noisy videos of other actions, or containing no action, in the support videos. Frame-level noise is included by not trimming the supports videos, thus several frames without common action support remain. The results are shown in Table~\ref{tab:noise}. When one out of five support videos contains the wrong action, the performance drops only 1.3\% from 28.1 to 26.8. Where we observe a noisy video containing no action is worse than a video containing a non-common action. When two noisy support video are from the same class, the drop is larger, which is to be expected, as this creates a stronger bias towards a distractor class. Adding a few noisy frames to the (25-frame) trimmed support videos results in slight performance drops, only when 32\% of the frames in each support video are noisy, results start to suffer a bit, with a drop from 28.1 to 24.5. Overall, we find that our approach is robust to both video-level and frame-level noise for common action localization in time and space.

\begin{figure*}
	\centering
	\includegraphics[width=0.96\linewidth]{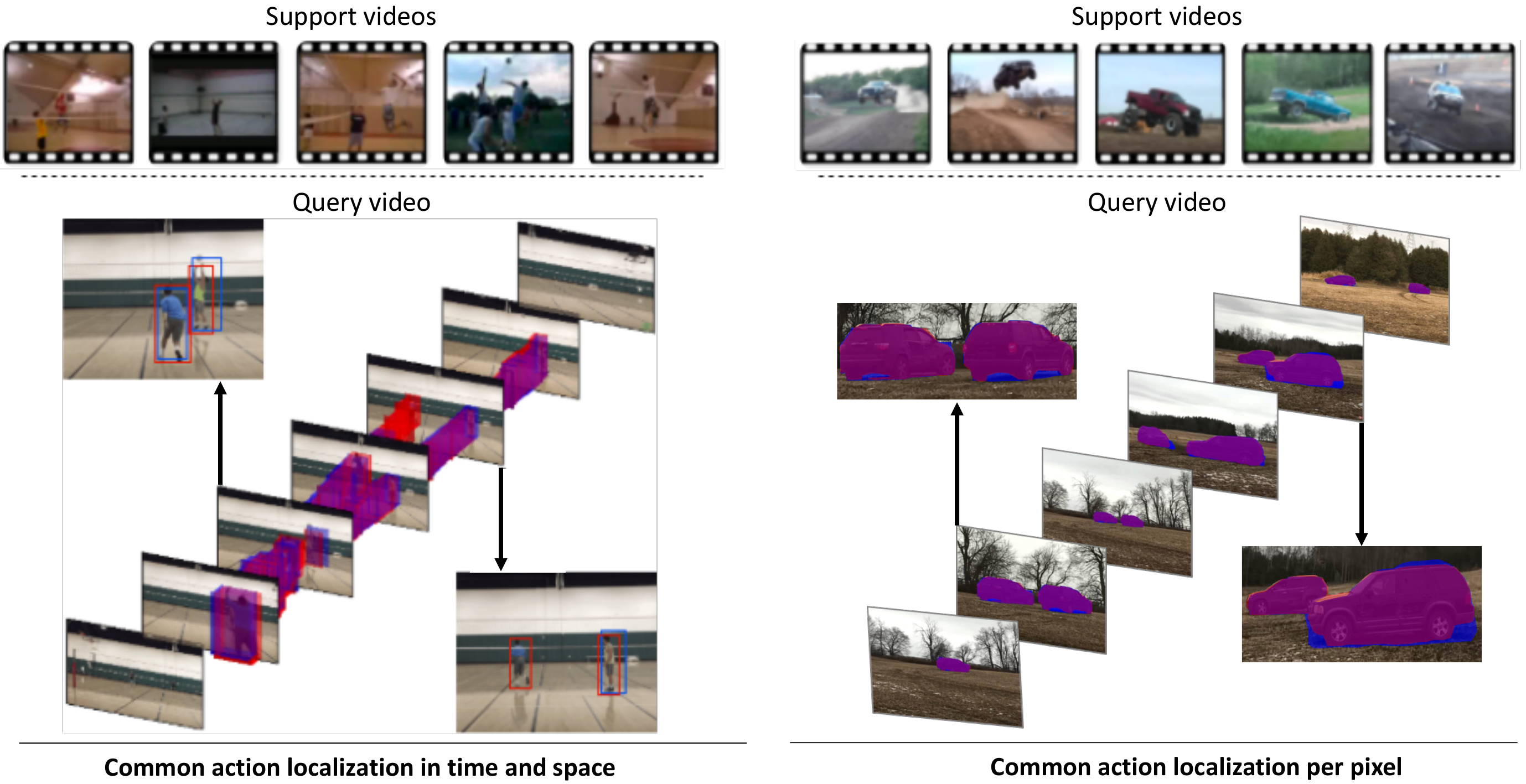}
	\caption{\textbf{Qualitative result} under one-shot (\textcolor{blue}{blue}) and five-shot (\textcolor{red}{red}) settings. In the upper part are 2 sets of 5 support videos, where the leftmost video in each set is also used in the one-shot setting. For the left example of common action localization in space and time, with one support video we only miss a short piece at the end of a common action tube. When more support videos are added, we recover the missed clip. For the right example of common action localization per pixel, with one support video we can localize the common action per pixel, and we present more precise localization per pixel with five support videos.}
	%\psmm{I like the 3d component. It does feel a bit like legos right not, the alignment between the support and query examples in both figures is almost "too perfect". Maybe make the bottom query part in both cases a bit smaller and add text "Support videos" "Query video" text etc in the figure.}~\pengwan{Maybe the alignment between query and support is not a problem. I just select the clearest frame in each support video where the subjects are most obvious.}}
	\label{fig:qualitative results}
\end{figure*}

\begin{table}
	\centering
	\resizebox{0.85\columnwidth}{!}{%
	\begin{tabular}{lc}
			\toprule
			\textbf{No noise} & 28.1\\
			\midrule
			\rowcolor{mygray}
			\textbf{Video-level noise} & \\
			1 noisy support video of other class & 26.8\\
			1 noisy support video without action & 26.3\\
			2 noisy support videos of different class & 25.3\\
			2 noisy support videos of same class & 24.7\\
			\bottomrule
			\rowcolor{mygray}
			\textbf{Frame-level noise} & \\
			2 noisy frames in each support video & 27.9\\
			4 noisy frames in each support video & 27.4\\
			6 noisy frames in each support video & 26.1\\
			8 noisy frames in each support video & 24.5\\
			\bottomrule
		\end{tabular}%
		}
		\caption{\textbf{Effect of noisy support videos} on Common-AVA for the five-shot setting. The result shows our robustness.}
		\label{tab:noise}
\end{table}

\begin{table*}[t!]
	\centering
	\resizebox{1.4\columnwidth}{!}{%
	\begin{threeparttable}
		\begin{tabular}{lrrrrrr}
			\toprule
			&\multicolumn{2}{c}{\textbf{Common-AVA}}&\multicolumn{4}{c}{\textbf{Common-UCF}}\\
		\cmidrule(lr){2-3} \cmidrule(lr){4-7}	&\multicolumn{2}{c}{frame-mAP}&\multicolumn{2}{c}{frame-mAP}&\multicolumn{2}{c}{video-mAP}\\
			& one-shot & five-shot & one-shot & five-shot & one-shot & five-shot\\
			\midrule
			Hu~\etal~\cite{hu2019silco} & 18.4 & 21.1 & 52.6 & 54.9 & 40.7 & 42.9\\
			Kang~\etal~\cite{kang2019few} & 19.7 & 21.5 & 54.6 & 56.2 & 42.5 & 43.8\\
			Feng \etal~\cite{feng2019spatio} & 21.6 & 23.2 & 59.4 & 61.3 & 46.9 & 48.5\\
			\textit{\textbf{This paper}} & \textbf{25.3} & \textbf{28.1} & \textbf{64.3} & \textbf{66.7} & \textbf{49.8} & \textbf{52.7}\\
			\bottomrule
		\end{tabular}%
% 		\footnotesize{$^\dagger$~We  adapted the one-stage object detector of Kang~\etal~\cite{kang2019few} to common spatio-temporal video localization.}\\
% 		\footnotesize{$^\ddagger$~We combined elements of the one-shot architecture of Feng \etal~\cite{feng2019spatio} and the few-shot ability of Yang \etal~\cite{yang2020localizing}.}\\ 
		\end{threeparttable}
	}
	\caption{\textbf{Common action localization in time and space} on Common-AVA and Common-UCF. We constructed two one-stage~baselines~\cite{hu2019silco,kang2019few} and a two-stage~baseline~\cite{feng2019spatio} from author-provided code.
	Our few-shot transformer obtains favourable results under the one- and five- shot settings on both datasets, without the need for proposals or other pre-defined components.} 
	\label{tab:spatio-temporal common localization}
\end{table*}

\begin{table}
	\centering
	\resizebox{0.99\columnwidth}{!}{%
		\begin{tabular}{lrrrr}
			\toprule
		&\multicolumn{2}{c}{\textbf{Common-Instance}}&\multicolumn{2}{c}{\textbf{Common-Multi-instance}}\\
			\cmidrule(lr){2-3} \cmidrule(lr){4-5}
			& one-shot & five-shot & one-shot & five-shot\\
			\midrule
		Feng~\etal~\cite{feng2018video} & 43.5 & n.a. & 31.4 & n.a.\\
		Zhang~\etal~\cite{zhang2020metal} & 49.5 & 52.0 & 41.4 & 43.5\\
		Yang~\etal~\cite{yang2020localizing} & 53.1 & 56.5 & 42.1 & 43.9\\
			\textit{\textbf{This paper}} (C3D) & 57.5 & 60.6 & 47.8 & 48.7\\
			\midrule
			\textit{\textbf{This paper}} (I3D) & \textbf{59.3} & \textbf{61.9} & \textbf{50.2} & \textbf{52.3}\\	
			\bottomrule
		\end{tabular}%
	}
	\caption{\textbf{Common action localization in time only.} All methods follow the setups of Feng~\etal~\cite{feng2018video} and Yang~\etal~\cite{yang2020localizing} on their ActivityNet derived datasets. The metric is video-mAP with an overlap threshold of 0.5. With the same C3D backbone we already outperform all methods, even though our approach is not designed for this setting.}
	\label{tab:temporal on activitynet}
\end{table}

\textbf{Common action localization in time and space.} In this experiment we demonstrate the effectiveness of our transformer model on the task of common action localization in time and space. As the task is novel, we cannot compare with existing methods that are not designed for this task. Instead, we have adapted existing one-stage and two-stage methods, intended for other tasks. Hu~\etal~\cite{hu2019silco} and Kang~\etal~\cite{kang2019few} provide one-stage approachs for few-shot common object detection. We replace their image feature extractors with an I3D network~\cite{carreira2017quo} to extract video features and transform the 4D video feature to a 3D feature by flattening along the temporal dimension and the channel dimension. With the flattened features, the few-shot object detector pipelines can localize the bounding box in the keyframe of each query video clip.
% This is our first ‘one-stage baseline’.
The spatio-temporal video relocalization of Feng~\etal~\cite{feng2019spatio} is a two-stage approach based on Faster R-CNN~\cite{ren2015faster}, which relies on a large set of proposals and contains several pre-defined components, such as the anchors in the proposal generation and non-maximum suppression in post-processing. Our few-shot transformer frees the need for proposals and such hand-crafted components. Also, where Feng~\etal~\cite{feng2019spatio} is restricted to the one-shot setting, we attempt to make their method compatible with our multi-shot setting. We do so by replacing the matching module of Feng~\etal~\cite{feng2019spatio} with the modules of Yang~\etal~\cite{yang2020localizing}, which can efficiently fuse multiple support videos with the query video. 
% This defines the second `two-stage baseline'. 

We compare to both baselines under the one- and five-shot setting on Common-AVA and Common-UCF in Table~\ref{tab:spatio-temporal common localization}. Our approach achieves the best result on both datasets, for both settings. Notably, our performance gains increase more than the baselines as more support videos are available, indicating that our transformer architecture is better able to leverage the commonalities between query and support features. We show  a qualitative result in Figure~\ref{fig:qualitative results} and provide more in the supplemental material. 

\textbf{Common action localization in time only.}
This task strives to localize the temporal extent of a common action in the untrimmed query video~\cite{yang2020localizing}. This is a related but different task from ours. Yet, our approach is naturally extended by changing the prediction outputs from a spatial bounding box for a query video clip to a temporal bounding box for the whole query video. We compare to  Yang~\etal~\cite{yang2020localizing}, Feng~\etal~\cite{feng2018video} and Zhang~\etal~\cite{zhang2020metal}. Yang~\etal~\cite{yang2020localizing} introduce a two-stage approach that localizes the common action in time from a few support videos, while Feng~\etal~\cite{feng2018video} introduce a one-stage approach for this task intended for a single support video. Zhang~\etal~\cite{zhang2020metal} perform few-shot temporal action localization, where video-level annotations are needed. When we assume the query and support videos share common actions, the one-way few-shot variant of Zhang~\etal becomes suitable for common action localization in time. We experiment on the reorganized datasets of Yang~\etal~\cite{yang2020localizing}, built upon Activity1.3~\cite{caba2015activitynet}, where the common instance split contains one single action in each video and the common multi-instance split contains multiple actions in each video. Results in Table~\ref{tab:temporal on activitynet} show 
results, where all methods use the same C3D~\cite{tran2015learning} backbone. We outperform all three methods on the common temporal localization task, even though our approach is not designed for this setting.

\textbf{Common action localization per pixel.} Finally, we demonstrate that our few-shot transformer is also suitable for common action localization per pixel. We consider this specialization of our task to further demonstrate the potential and generalization of our few-shot transformer. We simply add a mask-head on top of the decoder outputs, which predicts a binary mask for each pixel inside the predicted boxes. For evaluation we reorganize the A2D dataset by Xu \etal~\cite{xu2015can}, which provides dense pixel-level annotations for 3,782 videos of 43 actor-action classes. Among the 43 actor-action classes, 33 are used for training, 5 for validation and 5 for testing. The structure of our mask-head, the reorganization of Common-A2D, and the training details for common action localization per pixel are provided in the supplementary material. For comparison we adapt the few-shot common object segmentation approach by
Siam~\etal~\cite{siam2020weakly} as our baseline. We replace their flattened image feature sequences with the flattened video feature sequences. We compare to the baseline under one- and five- shot settings on Common-A2D in Table~\ref{tab:segmentation}. Our approach performs better under both the one- and five- shot settings. In Figure~\ref{fig:qualitative results} we also show a per-pixel segmentation result.
%\cs{plus sentence referring to figure}

\begin{table}
	\centering
	\resizebox{0.65\columnwidth}{!}{%
		\begin{tabular}{lrr}
			\toprule
				&\multicolumn{2}{c}{\textbf{Common-A2D}}\\
				\cmidrule(lr){2-3}
			& one-shot & five-shot\\
			\midrule
		Siam~\etal~\cite{siam2020weakly} & 43.3 & 44.8\\
			\textit{\textbf{This paper}} & \textbf{50.6} & \textbf{52.5}\\
			\bottomrule
		\end{tabular}%
	}
	\caption{\textbf{Common action localization per pixel} on reorganized A2D dataset. We adapted the few-shot object segmentation of Siam~\etal~\cite{siam2020weakly} to our task. The metric is mIoU across query videos. We perform better across both settings.}
	\label{tab:segmentation}
\end{table}
%=============================================
\section{Conclusion}
%=============================================
We consider spatio-temporal action localization in an untrimmed query video given a few trimmed support videos with a common action, without specifying the action label,
temporal bounds,
%or
%knowing
%the temporal
or spatial bounds. To tackle this challenge, we propose a few-shot transformer with a dedicated encoder-decoder structure optimized for joint commonality learning and localization prediction.
%, without the need for proposals.
Evaluation on reorganizations of AVA and UCF101-24 show that our approach localizes a common action in time and space, even when support videos are noisy. Although designed for spatio-temporal localization, our approach generalizes to temporal and per-pixel common action localization, outperforming the respective state-of-the-arts.

\textbf{Acknowledgements}. We would like to thank Tao Hu and Shuo Chen for helpful discussions and feedback.
{\small
\bibliographystyle{ieee_fullname}
\bibliography{7_egbib}
}
\clearpage
\clearpage
\section{Detailed architecture}

\begin{figure}[tb!]
	\centering
	\includegraphics[width=0.8\columnwidth]{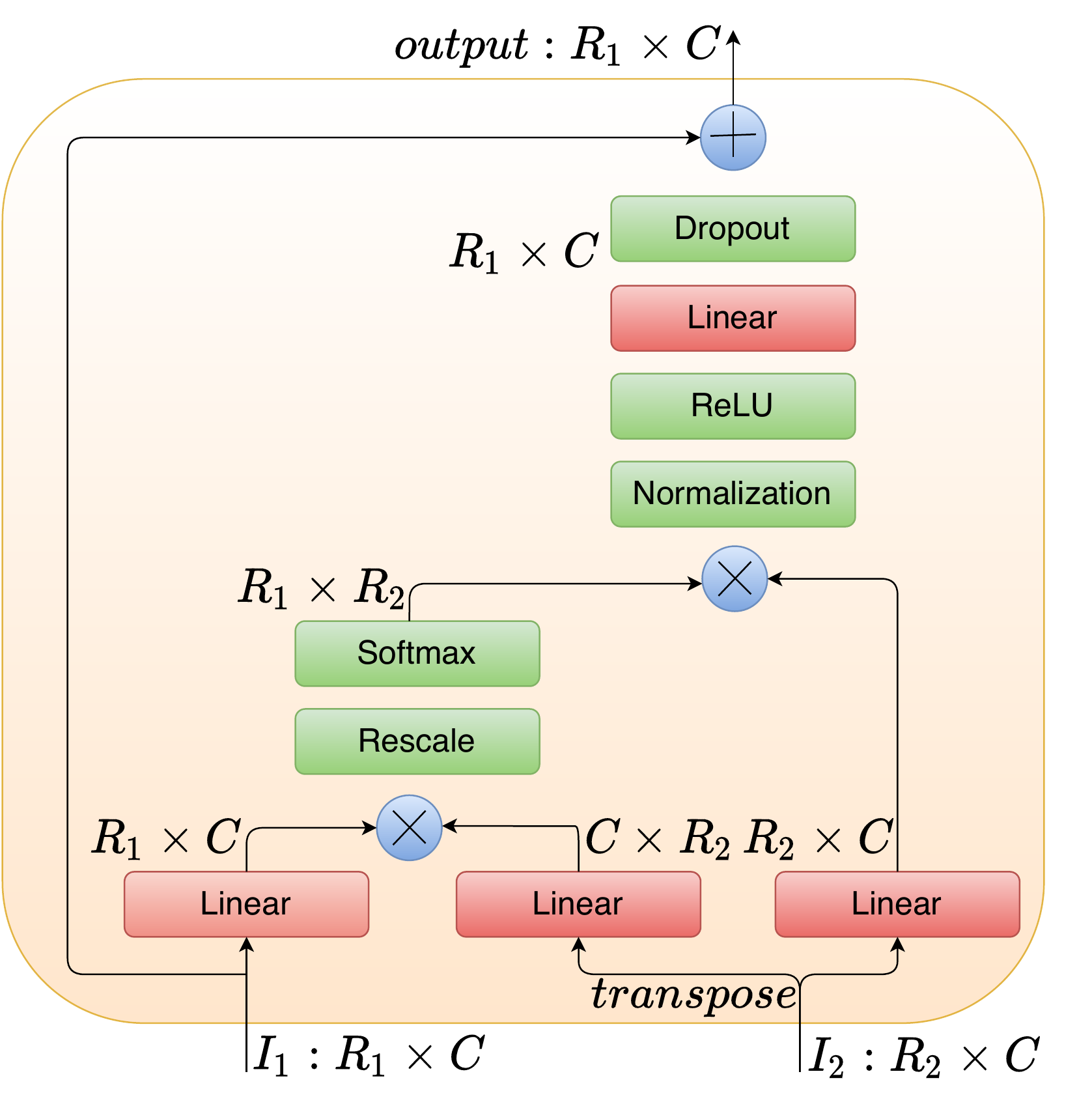}
	\caption{\textbf{Overview of common attention block.} $I_1$, $I_2$ denote the inputs of our common attention block. $\otimes$ denotes matrix multiplication and $\oplus$ is  element-wise sum. The main idea of the common attention block is to align the feature $I_2$ to the feature $I_1$.}
	\label{fig:common_block}
\end{figure}

\textbf{Overview of the common attention block.} The structure of the common attention block is illustrated in Figure~\ref{fig:common_block}. The main idea of the common attention block is to align the feature $I_2$ to the feature $I_1$. In our model, the common block plays two important roles: i) it aligns each query clip feature with its previous clip features to contain more motion information, ii) it fuses the support feature into the query clip feature based on the joint commonality.

\textbf{Spatio-temporal positional encoding.} In the encoder layers, both support and query branches are associated with corresponding spatio-temporal positions of video features. We generalize the original positional encoding~\cite{bello2019attention} to the 3D case. For all the spatio-temporal coordinates of each embedding, we independently use $ \frac{C}{3} $ sine and cosine functions with different frequencies. We then concatenate them to get the final $C$ channel positional encoding.

\begin{figure}[t]
\centering
\includegraphics[width=0.99\linewidth]{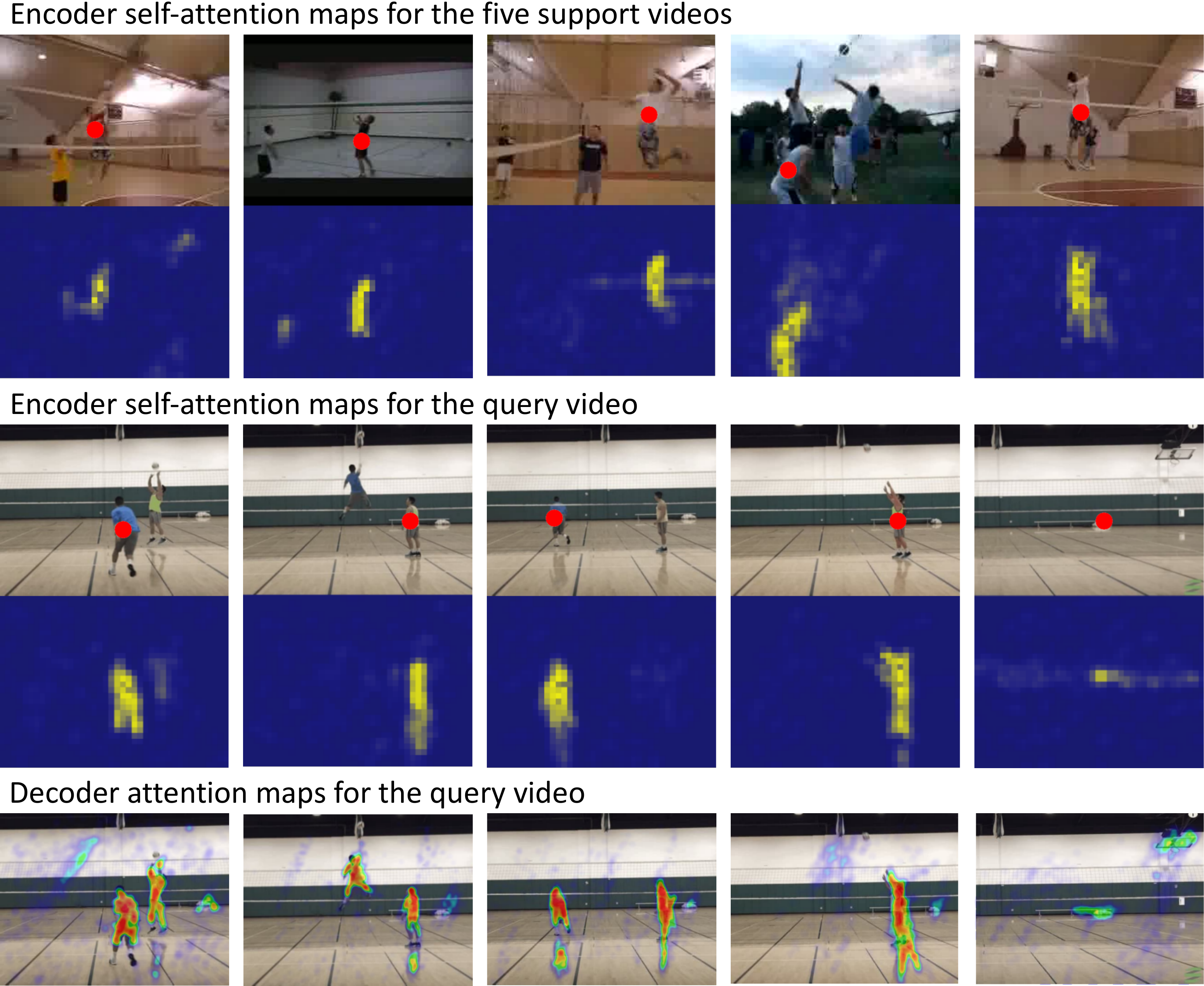}
  \caption{\textbf{Attention maps.} We show support and query encoder self-attention maps (top and middle), as well as query decoder attention maps (bottom). $\color{red} \bullet$ denotes reference point for self-attention map. The encoder makes individual actions stand out in the support and query videos. The decoder highlights the common actions in the query video.}
\label{fig:heatmap}
\end{figure}

\textbf{Visualization.} We visualize the attention maps in Figure~\ref{fig:heatmap} to better understand our model. The encoder self-attention maps are from the last encoder layer of a trained model. The decoder attention maps are the normalized attention score maps in the common attention block of the decoder. The figure shows that the encoder can make individual actions stand out in the support and query videos, which boosts commonality extraction for the decoder. On the basis of the encoder, the decoder is able to highlight the common actions in the query video.

\section{Additional ablations}

\begin{figure}[tb!]
	\centering
	\includegraphics[width=1\columnwidth]{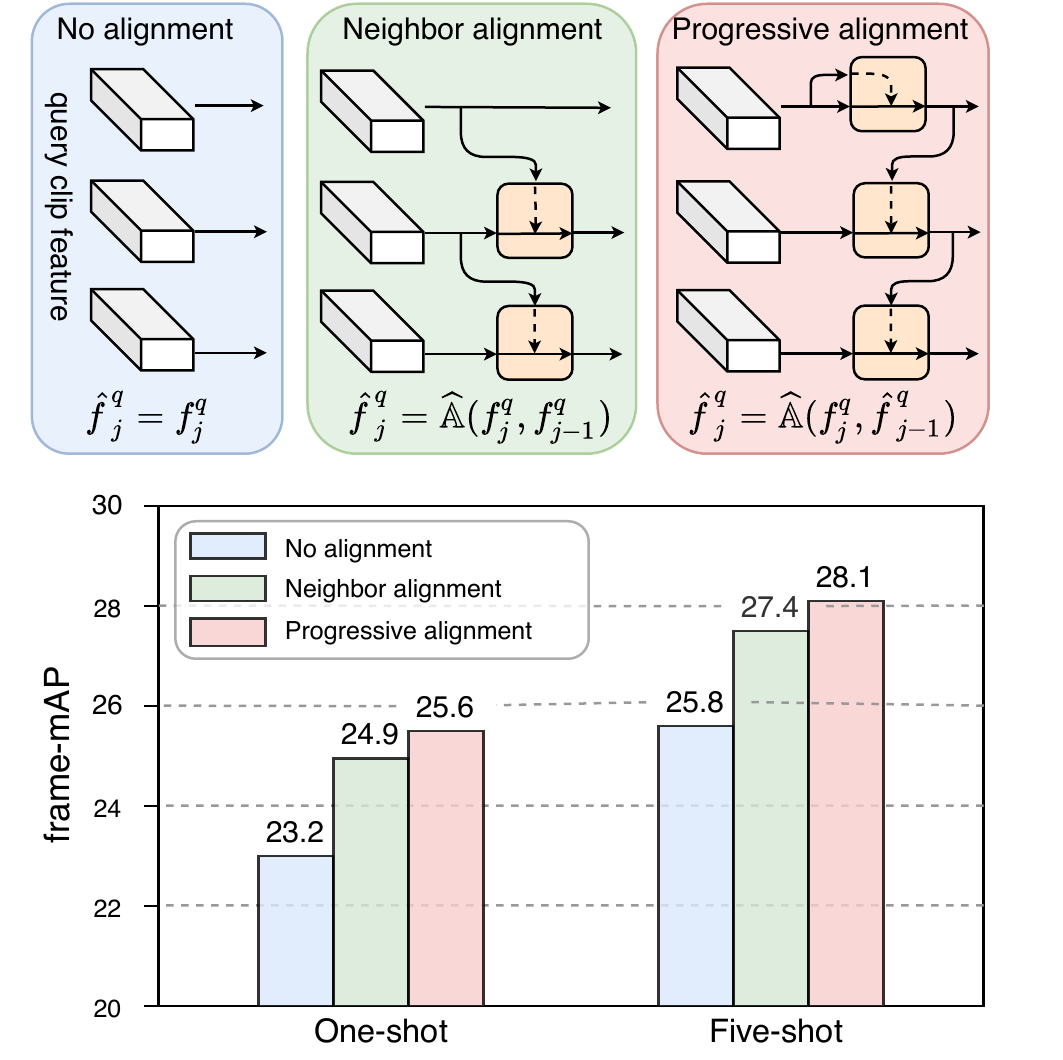}
	\caption{\textbf{Benefit of query clip feature alignment.} Propagating the spatio-temporal information from previous clips of the query video into the current query clip by using the common attention block is beneficial to the common localization on common-AVA. Progressively aligning previous query clip features into the current query clip leads to the best results.}
	\label{fig:query_alignment}
\end{figure}

\begin{figure*}
	\centering
	\includegraphics[width=0.99\linewidth]{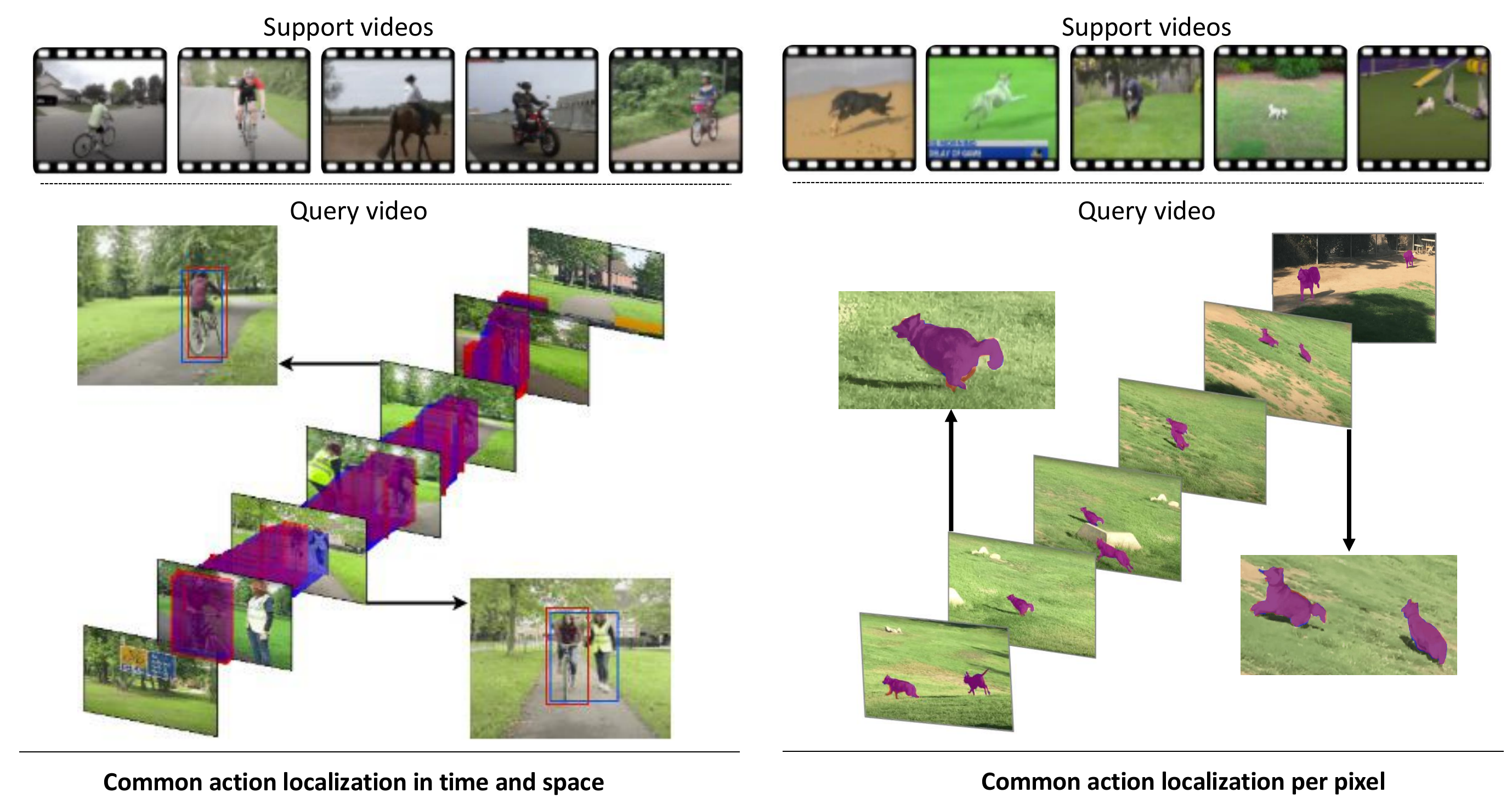}
	\caption{\textbf{Qualitative result} under one-shot (\textcolor{blue}{blue}) and five-shot (\textcolor{red}{red}) settings. In the upper part are 2 sets of 5 support videos, where the leftmost video in each set is also used in the one-shot setting. For the left example of common action localization in time and space, with one support video, we can find the common action tube in most clips of the query video, except for the few clips where we wrongly include an extra subject in our prediction. When we use five support videos, our bounding box is refined to exclude the redundant subject. For the right example of common action localization per pixel, with one support video we can localize the common action per pixel, and we present more precise localization with five support videos.}
	%\psmm{I like the 3d component. It does feel a bit like legos right not, the alignment between the support and query examples in both figures is almost "too perfect". Maybe make the bottom query part in both cases a bit smaller and add text "Support videos" "Query video" text etc in the figure.}~\pengwan{Maybe the alignment between query and support is not a problem. I just select the clearest frame in each support video where the subjects are most obvious.}}
	\label{fig:qualitative results}
\end{figure*}

\textbf{Benefit of query clip feature alignment.} We demonstrate the benefit of query clip feature alignment with the common attention block on the spatio-temporal localization performance on Common-AVA in Figure~\ref{fig:query_alignment}. 
%\cs{Cannot parse this sentence:}
%The neighbor alignment is to alignment the current query clip $f_j^q$ with its previous  neighbor query clip feature $f_{j-1}^q$ while the progressive alignment is to align with the previous enhanced clip feature $\hat{f}_{j-1}^q$. 
The neighbor alignment is aligns the current query clip with its \textit{single} previous neighbor query clip feature, while the progressive alignment aligns with \textit{all} previous clip features.
So the neighbor alignment lets each query clip contain the spatio-temporal information of its previous neighbor clip. And the progressive alignment propagates long-term motion information of previous clips to the current query clip. The neighbor alignment notably improves the performance and the progressive alignment causes a further  performance increase.

\textbf{Effect of variable-length support videos.} We verify our method can handle support videos of varying lengths in Table~\ref{tab:flexible_support}. This is indeed the case, especially when our model is also trained on videos of variable length.

\begin{table}
	\centering
	\resizebox{1\columnwidth}{!}{%
	\begin{tabular}{llc}
			\toprule
			\rowcolor{mygray}
			\textbf{Support videos in training} & \textbf{Support videos in evaluation} &\\
			\midrule
			\multirow{2}*{All videos are 5 frames}& All videos are 5 frames&22.2\\
			&The videos are 5,10,15,20,25 frames&23.8\\
			\midrule
			\multirow{2}*{All videos are 25 frames}& All videos are 25 frames&28.1\\
			&The videos are 5,10,15,20,25 frames&25.0\\
			\midrule
			The videos are 5,10,15,20,25 frames&The videos are 5,10,15,20,25 frames&26.1\\
			\bottomrule
		\end{tabular}%
		}
		\caption{
		\textbf{Effect of variable-length support videos} on five-shot Common-AVA. The results demonstrate our flexibility.}
		\label{tab:flexible_support}
\end{table}
\textbf{Qualitative results.} Some extra qualitative results for common action localization in time and space, and per pixel are shown in Figure~\ref{fig:qualitative results}.

\begin{figure*}[t!]
	\centering
	\includegraphics[width=0.99\linewidth]{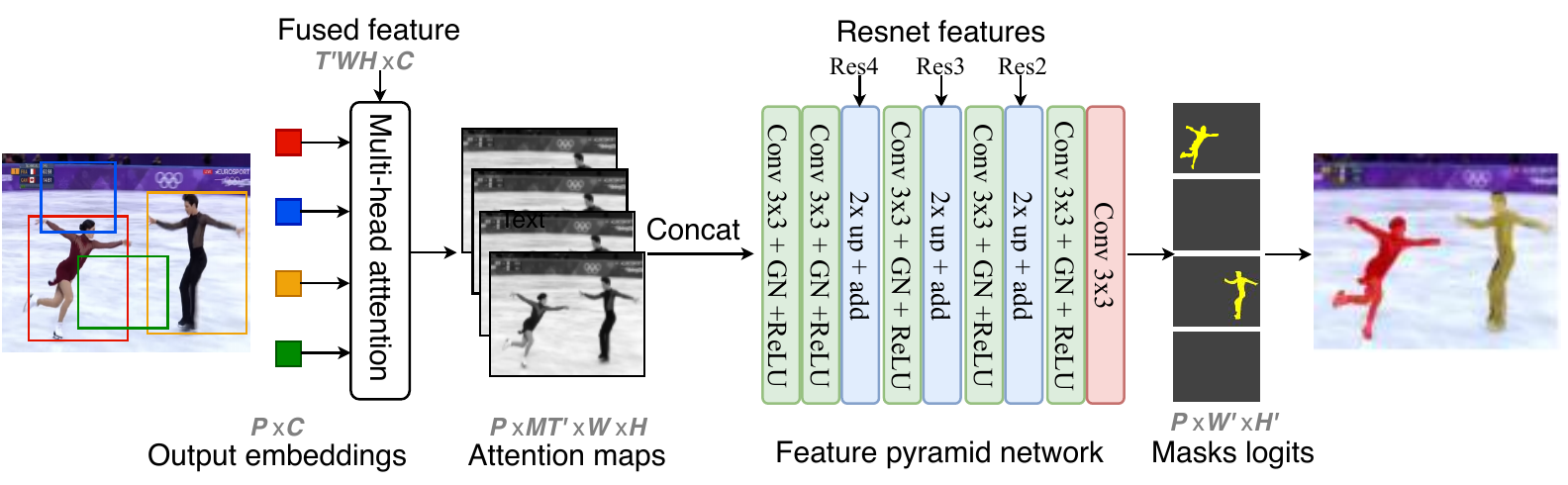}
	\caption{\textbf{Structure of the mask-head.} $M$ denotes the head number in the multi head attention, $W'$, $H'$ denote the width and height of the final masks. A binary mask is generated in parallel for each predicted box, then the masks are merged.}
	\label{fig:segmentation}
\end{figure*}

\section{Segmentation}

\textbf{The mask-head.} Inspired by the extension to segmentation in Carion~\etal~\cite{carion2020detr}, we localize the common action per pixel by simply adding a mask-head upon the decoder outputs, which predicts a binary mask for each of the predicted boxes, see Figure~\ref{fig:segmentation}. It takes as input the $output \  embeddings$ from the few-shot transformer decoder and computes multi-head attention weights of this embedding over the fused feature of the support and query branches from the encoder, generating attention maps per box in a small resolution. A feature pyramid network architecture~\cite{lin2017feature} is used to increase the resolution and make the final prediction with the supervision of DICE/F-1 loss~\cite{milletari2016fully} and Focal loss~\cite{lin2017focal}.

\textbf{Common-A2D.} The videos in the dataset have an average length of 136 frames where three to five frames for each video are labeled with dense pixel-level annotations. The selected frames are evenly distributed over a video. There are 2932 videos in the training subset, and 850 videos in the validation and testing subsets. For the training subset, we divide each query video into clips according to the labeled frames, to make each query clip contain one pixel-level annotated frame. Then we sample the query clips to a length of 25 frames. For the validation and testing subsets, we divide each query video into clips of 25 frames long without sampling.

\textbf{Training details.} The mask-head is trained jointly with the whole model for 100 epochs. During inference we first filter out the detection with a confidence below 85\% or background label, then compute the per-pixel argmax to determine whether each pixel is foreground.

\end{document}